\documentclass[10pt,journal,compsoc]{IEEEtran}

\usepackage{graphicx}
\usepackage[tight,normalsize,sf,SF]{subfigure}
\usepackage{booktabs} 
\usepackage{amsmath}
\usepackage{amssymb}
\usepackage{array}
\usepackage{multirow}
\usepackage{color}
\usepackage{colortbl}
\usepackage{ragged2e}
\def\mycolor{\cellcolor[rgb]{0.8275,0.8275,0.8275}}
\def\myTextColor{\textcolor[rgb]{0, 0, 0}}
\def\etal{\emph{et al}.}
\def\etc{\emph{etc}.}
\def\eg{\emph{e.g.}}
\def\ie{\emph{i.e.}}
\def\T{\mathrm{T}}
\usepackage[pagebackref=false,breaklinks=true,letterpaper=true,colorlinks,bookmarks=false]{hyperref}

\begin{document}
\title{Adversarial Metric Attack and Defense for Person Re-identification}
\author{Song Bai, Yingwei Li, Yuyin Zhou, Qizhu Li, and Philip H.S. Torr
\IEEEcompsocitemizethanks{\IEEEcompsocthanksitem S. Bai, Q. Li and P.H. Torr are with the University of Oxford, UK. E-mail: songbai.site@gmail.com, \{qizhu.li, philip.torr\}@eng.ox.ac.uk.
\IEEEcompsocthanksitem Y. Li and Y. Zhou are with the Johns Hopkins University, USA. E-mail: yingwei.li@jhu.edu, zhouyuyiner@gmail.com.}
}

\markboth{IEEE Transactions on Pattern Analysis and Machine Intelligence}%
{Shell \MakeLowercase{\textit{et al.}}: Bare Demo of IEEEtran.cls for Computer Society Journals}

\IEEEtitleabstractindextext{
\justify
    \begin{abstract}
    Person re-identification (re-ID) has attracted much attention recently due to its great importance in video surveillance. In general, distance metrics used to identify two person images are expected to be robust under various appearance changes. However, our work observes the extreme vulnerability of existing distance metrics to adversarial examples, generated by simply adding human-imperceptible perturbations to person images. Hence, the security danger is dramatically increased when deploying commercial re-ID systems in video surveillance. 
    Although adversarial examples have been extensively applied for classification analysis, it is rarely studied in metric analysis like person re-identification. The most likely reason is the natural gap between the training and testing of re-ID networks, that is, the predictions of a re-ID network cannot be directly used during testing without an effective metric. In this work, we bridge the gap by proposing Adversarial Metric Attack, a parallel methodology to adversarial classification attacks. Comprehensive experiments clearly reveal the adversarial effects in re-ID systems. Meanwhile, we also present an early attempt of training a metric-preserving network, thereby defending the metric against adversarial attacks. At last, by benchmarking various adversarial settings, we expect that our work can facilitate the development of adversarial attack and defense in metric-based applications.
    \end{abstract}
\begin{IEEEkeywords}
Person Re-identification, Adversarial Attack, Metric Learning.
\end{IEEEkeywords}
}
\maketitle
\IEEEdisplaynontitleabstractindextext
\IEEEpeerreviewmaketitle

\IEEEraisesectionheading{\section{Introduction}\label{sec:intro}}
\IEEEPARstart{I}n recent years, person re-identification (re-ID)~\cite{karanam2016systematic,zheng2016person} has attracted great attention in the computer vision community, driven by the increasing demand of video surveillance in public space. Hence, great effort has been devoted to developing robust re-ID features~\cite{symmetry,VIPeR_ELF,fu2018horizontal,li2019unsupervised} and distance metrics~\cite{RankSVM,PRDC,XQDA,Duan_2018_CVPR} to overcome the large intra-class variations of person images in viewpoint, pose, illumination, blur, occlusion and resolution. For example, the rank-1 accuracy of the latest state-of-the-art on the Market-1501 dataset~\cite{market1501} is $93.8$~\cite{HACNN}, increasing rapidly from $44.4$ when the dataset was first released in $2015$.

\begin{figure}[tb]
\centering
\includegraphics[width = 0.42\textwidth]{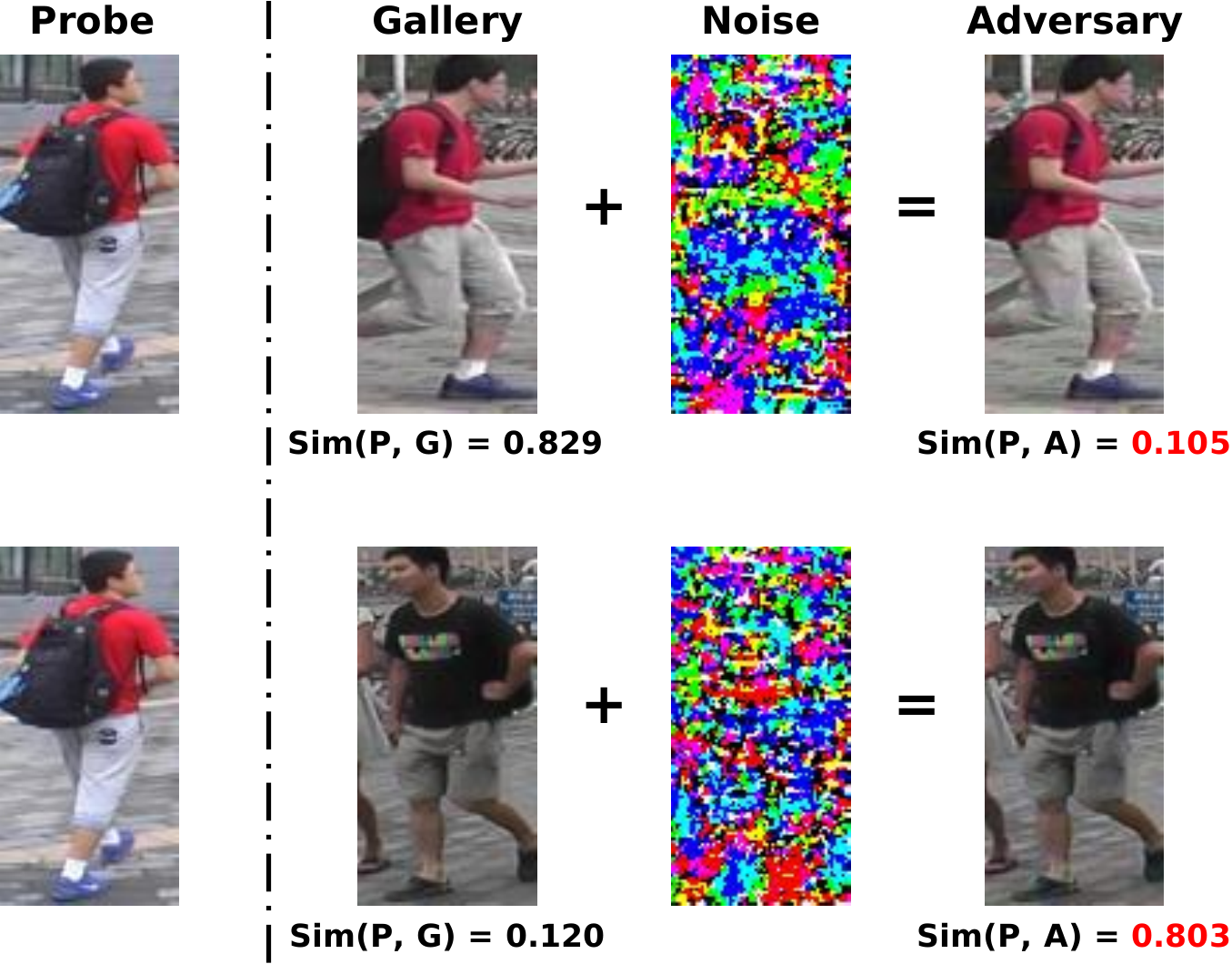}
\vspace{-2ex}
\caption{The illustration of the adversarial effect in person re-identification. Given a probe image, its similarity with the true positive is decreased from 0.829 to 0.105, and that with the true negative is increased from 0.120 to 0.803 by adding human-imperceptible noise to gallery images. The adversarial noise is resized to the range $[0,1]$ for visualization.}
\label{fig:ill_perturbation}
\vspace{-3ex}
\end{figure}

However, we draw researchers' attention to the fact that re-ID systems can be very vulnerable to adversarial attacks. Fig.~\ref{fig:ill_perturbation} shows a case where a probe image is presented. Of the two gallery images, the true positive has a large similarity value and the true negative has a small one. Nevertheless, after adding human-imperceptible perturbations to the gallery images, the metric is easily fooled even though the new gallery images appear the same as the original ones. In real applications, such vulnerability could be used in a malicious way (\eg, criminals hide themselves in a database) or a benign way (\eg, protecting people with sensitive identities from being tracked by a video surveillance system).

\begin{figure}[tb]
\centering
\subfigure[]
{
\begin{minipage}[tb]{0.22\textwidth}
\includegraphics[width = 1\textwidth]{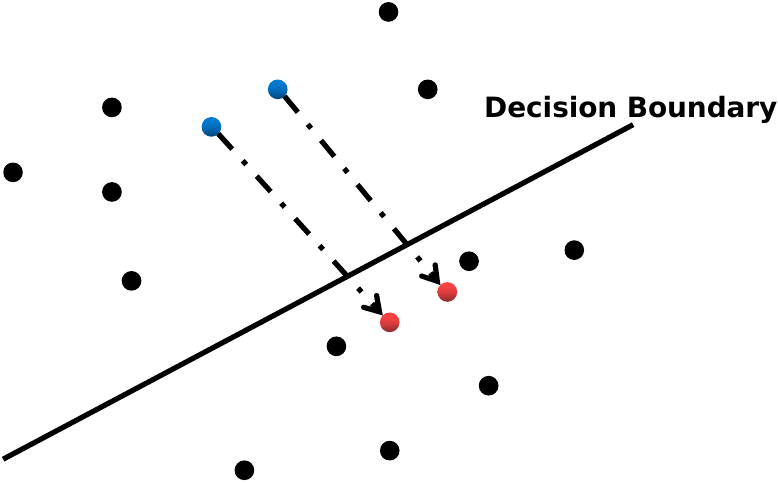}
\end{minipage}
}
\subfigure[]
{
\begin{minipage}[tb]{0.22\textwidth}
\includegraphics[width = 1\textwidth]{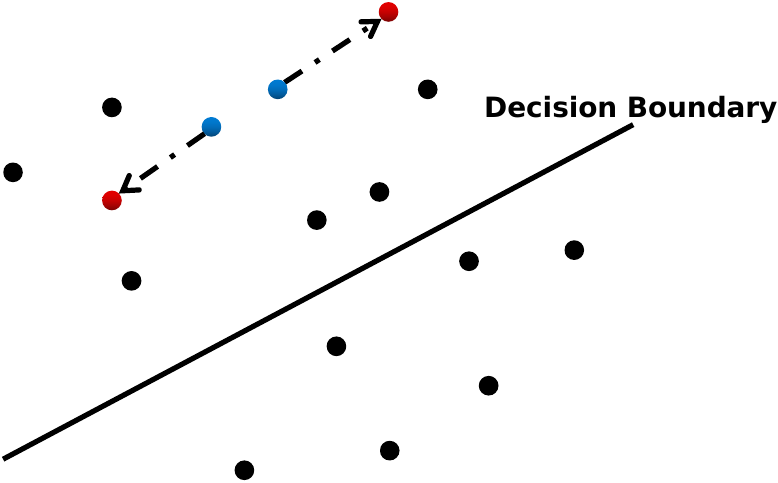}
\end{minipage}
}
\vspace{-2ex}
\caption{The failure case (a) of classification attack and the successful case (b) of metric attack on two clean images (blue color). The adversarial examples (red color) generated by the classification attack cross over the class decision boundary, but preserve the pairwise distance between them to a large extent.}
\label{fig:attack_cls}
\vspace{-2ex}
\end{figure}

Adversarial examples have been extensively investigated in classification analysis~\cite{I_FGSM,MI_FGSM}, such as image classification, object detection, semantic segmentation,~\etc~However, they have not attracted much attention in the field of re-ID, a metric analysis task whose basic goal is to learn a discriminative distance metric. A very likely reason is the existence of a natural gap between the training and testing of re-ID networks. While a re-ID model is usually trained with a certain classification loss, it discards the concept of class decision boundaries during testing and adopts a metric function to measure the pairwise distances between the probe and gallery images. Consequently, previous works on classification attacks~\cite{I_FGSM,MI_FGSM} do not generalize to re-ID systems,~\ie,~they attempt to push images across the class decision boundaries and do not necessarily lead to a corrupted pairwise distance between images (see Fig.~\ref{fig:attack_cls}).
Note that some re-ID networks are directly guided by metric losses (\eg,~contrastive loss~\cite{contrastive}), and their output can measure the between-object distances. However, it is still infeasible to directly attack such output owing to the sampling difficulty and computational complexity. Therefore, a common practice in re-ID is to take the trained model as a feature extractor and measure the similarities in a metric space. 

Considering the importance of security for re-ID systems and the lack of systematic studies on its robustness towards adversarial examples, we propose Adversarial Metric Attack, an efficient and generic methodology to generate adversarial examples by attacking metric learning systems. The contributions of this work can be divided into five folds

\vspace{1ex}\noindent\textbf{1)}~Our work presents what to our knowledge is the first systematic and rigorous investigation of adversarial effects in person re-identification, which should be taken into consideration when deploying re-ID algorithms in real surveillance systems.

\vspace{1ex}\noindent\textbf{2)}~We propose \emph{adversarial metric attack}, a parallel methodology to the existing \emph{adversarial classification attack}~\cite{szegedy2013intriguing,FGSM}, 
which can be potentially applied to other safety-critical applications that rely on distance metrics (\eg,~face verification~\cite{Deepface,kaziakhmedov2019real} and tracking~\cite{KCF}).

\vspace{1ex}\noindent\textbf{3)}~We define and benchmark various experimental settings for metric attack in re-ID, including white-box and blackbox attack, non-targeted and targeted attack, single-model and multi-model attack,~\etc~Under those experimental settings, comprehensive experiments are carried out with different distance metrics and attack methods.

\vspace{1ex}\noindent\textbf{4)}~We present an early attempt on adversarial metric defense, and show that adversarial examples generated by attacking metrics can be used in turn to train a metric-preserving network.

\vspace{1ex}\noindent\textbf{5)}~The code is publicly available to easily generate the \emph{adversarial version of benchmark datasets} (see examples in the supplementary material), which can serve as a useful testbed to evaluate the robustness of re-ID algorithms.

We hope that our work can facilitate the development of robust feature learning and accelerate the progress on adversarial attack and defense of re-ID systems with the methodology and the experimental conclusions presented. 

\section{Related Work}
Adversarial learning~\cite{huang2018adversarially,wang2018cascaded,DBLP:conf/ijcai/YinZWYWGHL18,qian2018pose} has been incorporated into the training procedure of re-ID systems in many previous works. In these works, generative adversarial networks (GAN)~\cite{GAN} typically acts as a data augmentation strategy by generating photo-realistic person images to enhance the training set. For example, Zheng~\etal~\cite{duke} apply GANs to generate unlabeled images and assigned a uniform label distribution during training. Wei~\etal~\cite{Wei_2018_CVPR} propose Person Transfer Generative Adversarial Network (PTGAN) to bridge the gap between different datasets. Moreover, Ge~\etal~\cite{ge2018fd}~propose Feature Distilling Generative Adversarial Network to learn identity-related and pose-unrelated representations. Adversarial PersonNet~\cite{li2018adversarial} uses GANs to address open-world person re-identification, a different setting as opposed to the closed-set conventional re-ID problem. It defines ``attack" as a process of using generative adversarial networks to generate target-like imposters for training the discriminator and defines ``defense" as a process of separating imposter samples from target classes. In~\cite{liu2018adversarial}, binary codes are learned for efficient pedestrian matching via Adversarial Binary Coding.

However, to the best of our knowledge, no prior work has rigorously considered the robustness of re-ID systems towards adversarial attacks, which have received wide attention in the context of classification-based tasks, including image classification~\cite{I_FGSM}, object detection and semantic segmentation. As these vision tasks aims to sort an \textit{instance} into a \textit{category}, they are therefore special cases of the broader classification problem.
On such systems, it has been demonstrated that adding carefully generated human-imperceptible perturbations to an input image can easily cause the network to misclassify the perturbed image with high confidence. These tampered images are known as adversarial examples and great efforts have been devoted to the generation of adversarial examples~\cite{FGSM,I_FGSM,MI_FGSM}. 
In contrast, our work focuses on adversarial attacks on metric learning systems, which analyze the relationship between two \textit{instances}. A relevant work to ours is~\cite{kaziakhmedov2019real}, where a real-world attack system is proposed for face detection.

\section{Adversarial Metric Attack} \label{sec:methodology}
Person re-identification~\cite{gong2014person} is comprised of three sets of images, including the probe set $\mathbf{P}=\{p_i\}_{i=1}^{N_p}$, the gallery set $\mathbf{X}=\{x_i\}_{i=1}^{N_x}$, and the training set $\mathbf{Y}=\{y_i\}_{i=1}^{N_y}$. A label set $\mathbf{L}$ is also given to annotate the identity of each image for training and evaluation. A general re-ID pipeline is: 1) learn a feature extractor $\mathbf{F}$ with parameters $\boldsymbol{\Theta}$ (usually by training a neural network) by imposing a loss function $\mathbf{J}$ to $\mathbf{L}$ and $\mathbf{Y}$; 2) extract the activations of intermediate layers for $\mathbf{P}$ and $\mathbf{X}$ as their visual features $\mathbf{F(P,\boldsymbol{\Theta})}$ and $\mathbf{F(X,\boldsymbol{\Theta})}$, respectively; 3) compute the distance between $\mathbf{F(P,\boldsymbol{\Theta})}$ and $\mathbf{F(X,\boldsymbol{\Theta})}$ for indexing. When representing features, $\mathbf{F(\cdot)}$ and $\boldsymbol{\Theta}$ are omitted where possible for notation simplicity.





In this paper, we aim to generate adversarial examples for re-ID models. As explained in Sec.~\ref{sec:intro}, a different attack mechanism is required for metric learning systems as opposed to the existing attack methods which focus on classification-based models~\cite{FGSM,I_FGSM}. Instead of attacking the loss function used in training the neural network as done in these previous works, we discard the training loss and propose to attack the distance metric. Such an attack mechanism directly results in the corruption of the pairwise distance between images, thus leading to guaranteed accuracy compromises of a re-ID system. This is the gist of the methodology proposed in this work, which we call adversarial metric attack. 

Adversarial metric attack consists of four components, including models for attack, metrics for attack, methods for attack and adversarial settings for attack. In the first component (Sec.~\ref{sec:dfl}), we train the model (with parameters $\boldsymbol{\Theta}$) on the training set $\mathbf{Y}$ as existing re-ID algorithms do. The model parameters are then fixed during attacking. In the second component (Sec.~\ref{sec:MFA}), a metric loss $\textbf{D}$ is determined as the attack target. In the third component (Sec.~\ref{sec:attack_methods}), an optimization method for producing adversarial examples is selected. In the last component (Sec.~\ref{sec:bs}), by setting the probe set $\mathbf{P}$ as a reference, we generate adversarial examples on the gallery set $\mathbf{X}$ in a specific adversarial setting.

\subsection{Models for Attack} \label{sec:dfl}
In the proposed methodology, the model for attack is not limited to be classification-based as opposed to~\cite{FGSM,I_FGSM}. Instead, most re-ID models~\cite{defense_triplet,quadruplet,Song_2018_CVPR,Liu_2018_CVPR} can be used. We only review two representative baseline models, which are commonly seen in person re-identification.

\vspace{1ex}\noindent\textbf{Cross Entropy Loss.}~The re-ID model is trained with the standard cross-entropy loss and the labels are the identities of training images. It is defined as
\begin{equation}
\mathbf{J}(\mathbf{Y},\mathbf{L})=-\sum_i\sum_j{\mathbf{1}\left(l(y_i)=j\right)\log q_i^j},
\end{equation}
where $q_i^j$ is the classification probability of the $i$-th training sample to the $j$-th category and $l(y_i)$ is the ground-truth label of $y_i\in\mathbf{Y}$. 

\vspace{1ex}\noindent\textbf{Triplet Loss.}~The re-ID model is trained with the triplet loss, defined as 
\begin{equation}
\mathbf{J}(\mathbf{Y},\mathbf{L})=\sum_{l_a=l_p\neq l_n}[d(y_a,y_p)-d(y_a,y_n)+m]_+,
\end{equation}
where $y_a$ denotes the anchor point, $y_p$ denotes the positive point and $y_n$ denotes the negative point. The motivation is that the positive $y_p$ belonging to the same identity as the anchor $y_a$ is closer to $y_a$ than the negative $y_n$ belonging to another identity, by at least a margin $m$.

\subsection{Metrics for Attack} \label{sec:MFA}
Metric learning (\eg,~XQDA~\cite{XQDA}, KISSME~\cite{KISSME}) has dominated the landscape of re-ID for a long time. Mathematically, a metric defined between the probe set $\mathbf{P}$ and the gallery set $\mathbf{X}$ is a function $\mathbf{D}: \mathbf{P}\times\mathbf{X}\rightarrow[0,\infty)$,
which assigns non-negative values for each pair of $p\in\mathbf{P}$ and $x\in\mathbf{X}$. We also use notation $d(p,x)$ to denote the distance between $p$ and $x$ in the metric space $\mathbf{D}$. 
In this section, we give the formal definition of metric loss used in adversarial metric attack. It should be mentioned that any differentiable metric (or similarity) function can be used as the target loss. 

\vspace{1ex}\noindent\textbf{Euclidean distance} is a widely used distance metric. The metric loss is defined as
\begin{equation} \label{eq:euclidean}
d(p,x)=\|p-x\|^2_2,
\end{equation}
which computes the squared Euclidean distance between $p$ and $x$.

\vspace{1ex}\noindent\textbf{Mahalanobis distance} is a generalization of the Euclidean distance that considers the correlation of different feature dimensions. Accordingly, we can have a metric loss as
\begin{equation} \label{eq:xqda}
d(p,x)=(p-x)^{\T}\mathbf{M}(p-x),
\end{equation}
where $\mathbf{M}$ is a positive semidefinite matrix.

\subsection{Methods for Attack} \label{sec:attack_methods}
Given a metric loss defined above, we aim at learning an adversarial example $x^\text{adv}=x+r$, where $x\in\mathbf{X}$ denotes a certain gallery image and $r$ denotes the adversarial perturbation. $L_\infty$ norm is used to measure the perceptibility of the perturbation,~\ie,~$\|r\|_\infty\leq\epsilon$ and $\epsilon$ is a small constant.

To this end, we introduce the following three attack methods, including:

\vspace{1ex}\noindent\textbf{Fast Gradient Sign Method (FGSM)~\cite{FGSM}} is a single step attack method. It generates adversarial examples by
\begin{equation} \label{eq:FGSM} 
\mathbf{X}^{\text{adv}}=
\mathbf{X}+\epsilon\cdot\text{sign}\left(\frac{\partial\mathbf{\mathbf{D}(\mathbf{P},\mathbf{X})}}{\partial\mathbf{X}}\right),
\end{equation}
where $\epsilon$ measures the maximum magnitude of adversarial perturbation and $\text{sign}(\cdot)$ denotes the signum function.

\vspace{1ex}\noindent\textbf{Iterative Fast Gradient Sign Method (I-FGSM)~\cite{I_FGSM}} is an iterative version of FGSM, defined as 
\begin{equation} \label{eq:I-FGSM}
\begin{cases}
\mathbf{X}_{0}^{\text{adv}}=\mathbf{X} \\
\mathbf{X}_{n+1}^{\text{adv}}=\boldsymbol{\Psi}_\mathbf{X}^{\epsilon} \left(\mathbf{X}_{n}^{\text{adv}}+\alpha\cdot\text{sign}(\frac{\partial\mathbf{D}(\mathbf{P},\mathbf{X}_{n}^{\text{adv}})}{\partial\mathbf{X}_{n}^{\text{adv}}})\right),
\end{cases}
\end{equation}
where $n$ denotes the iteration number and $\alpha$ is the step size. $\boldsymbol{\Psi}_\mathbf{X}^{\epsilon}$ is a clip function that ensures the generated adversarial example is within the $\epsilon$-ball of the original image.

\vspace{1ex}\noindent\textbf{Momentum Iterative Fast Gradient Sign Method (MI-FGSM)}~\cite{MI_FGSM} adds the momentum term on top of I-FGSM to stabilize update directions. It is defined as
\begin{equation} \label{eq:MI-FGSM}
\begin{cases}
g_{n+1}=\mu\cdot g_n+\frac{\partial\mathbf{D}(\mathbf{P},\mathbf{X}_{n}^{\text{adv}})}{\partial\mathbf{X}_{n}^{\text{adv}}}\bigg/\bigg\|\frac{\partial\mathbf{D}(\mathbf{P},\mathbf{X}_{n}^{\text{adv}})}{\partial\mathbf{X}_{n}^{\text{adv}}}\bigg\|_1\\
\mathbf{X}_{n+1}^{\text{adv}}=\boldsymbol{\Psi}_\mathbf{X}^{\epsilon} \left(\mathbf{X}_{n}^{\text{adv}}+\alpha\cdot\text{sign}(g_{n+1})\right),
\end{cases}
\end{equation}
where $\mu$ is the decay factor of the momentum term and $g_n$ is the accumulated gradient at the $n$-th iteration.

\subsection{Benchmark Adversarial Settings} \label{sec:bs}
In this section, we benchmark the experimental settings for adversarial metric attack in re-ID .

\subsubsection{White-box and Black-box Attack} 
\noindent\textbf{White-box attack} requires the attackers to have prior knowledge of the target networks, which means that the adversarial examples are generated with and tested on the same network having parameters $\mathbf{\Theta}$. It should be mentioned that for adversarial metric attack, the loss layer used during training is replaced by the metric loss when attacking the network.

\vspace{1ex}\noindent\textbf{Black-box attack} means that the attackers do not know the structures or the weights of the target network. That is to say, the adversarial examples are generated with a network having parameters $\mathbf{\Theta}$ and used to attack metric on another network which differs in structures, parameters or both.

\subsubsection{Targeted and Non-targeted Attack}
\noindent\textbf{Non-targeted attack} aims to widen the metric distance between image pairs of the same identity. Given a probe image $p$ and a gallery image $x$, where $l(p)=l(x)$, their distance $d(p, x)$ is ideally small. After imposing a non-targeted attack on the distance metric, the distance $d(p, x^{\text{adv}})$ between $p$ and the generated adversarial example $x^\text{adv}$ is enlarged. Hence, when $p$ serves as the query, $x^\text{adv}$ will not be ranked high in the ranking list of $p$ (see Fig.~\ref{fig:non-targeted}).

Non-targeted attack can be achieved by applying the attack methods described in Sec.~\ref{sec:attack_methods} to the metric losses described in Sec.~\ref{sec:MFA}.

\vspace{1ex}\noindent\textbf{Targeted attack} aims to draw the gallery image towards the probe image in the metric space. This type of attack is usually performed on image pairs with different identities,~\ie,~ $l(p)\neq l(x)$, which correspond to a large $d(p, x)$ value. The generated $x^\text{adv}$ becomes closer to the query image $p$ in the metric space, deceiving the network into predicting $l(x^\text{adv})=l(p)$. Hence, one can frequently observe adversarial examples generated by a targeted attack in top positions of the ranking list of $p$ (see Fig.~\ref{fig:targeted}). 

Unlike non-targeted attack where adversarial examples do not steer the network towards a specific identity, targeted attack finds adversarial perturbations with pre-determined target labels during the learning procedure and tries to decrease the value of objective function. This incurs a slight modification to the attack methods described in Sec.~\ref{sec:attack_methods}. For example, the formulation of FGSM~\cite{FGSM} is changed to
\begin{equation}
\mathbf{X}^{\text{adv}}=
\mathbf{X}-\epsilon\cdot\text{sign}\left(\frac{\partial\mathbf{D}(\mathbf{P},\mathbf{X})}{\partial\mathbf{X}}\right).
\end{equation}
The update procedure of I-FGSM~\cite{I_FGSM} and MI-FGSM~\cite{MI_FGSM} can be modified similarly.

\subsubsection{Single-model and Multi-model Attack}
\noindent\textbf{Single-model attack} differs from \textbf{Multi-model attack} in that the former only uses one network to learn the adversarial examples, while the latter uses multiple ones. It has been shown in the context of adversarial classification attacks~\cite{liu2016delving} that an ensemble of multiple models is crucial to the transferability of the adversarial examples. Thus, multi-model methods generally perform better under the black-box setting.

To ensemble multiple networks, we suggest to average the metric losses defined in Sec.~\ref{sec:MFA}. The logits or predictions of networks are not used in the multi-model metric attack, since they do not necessarily have the same dimension in contrast with the case in classification attack~\cite{MI_FGSM}.

\section{Adversarial Metric Defense}
Here we present an early attempt on training a metric-preserving network to defend a distance metric. 

The procedure is divided into four steps: 1) learn a clean model $\mathbf{F}$ with parameters $\boldsymbol{\Theta}$ by imposing a loss function $\mathbf{J}$ to $\mathbf{L}$ and $\mathbf{Y}$; 2) by treating each image in the training set $\mathbf{Y}$ as the query in turn, perform adversarial metric attack described in Sec.~\ref{sec:methodology} on $\mathbf{F}$, then obtain the adversarial version of training set $\mathbf{Y}^\text{adv}$; 3) merge $\mathbf{Y}$ and $\mathbf{Y}^\text{adv}$, and re-train a metric-preserving model $\mathbf{F}^\text{adv}$; 4) use $\mathbf{F}^\text{adv}$ as the testing model in replace of $\mathbf{F}$.

As for the performance, we find that $\mathbf{F}^\text{adv}$ closely matches $\mathbf{F}$ when testing on the original (clean) gallery set $\mathbf{X}$, but significantly outperforms $\mathbf{F}$ when testing on the adversarial version of gallery set $\mathbf{X}^\text{adv}$. In this sense, re-ID systems gain the robustness to adversarial metric attacks.

\section{Experiments}

\noindent\textbf{Datasets.}~Market-1501 dataset~\cite{market1501} is a widely accepted benchmark for person re-ID. It consists of $1501$ identities, in which $750$ identities ($12,936$ images) are used for training, $751$ identities ($19,732$ images) for testing and $3,368$ images for querying. DukeMTMC-reID dataset~\cite{duke} has $36,411$ images taken by $8$ cameras. The training set has $16,522$ images ($702$ identities). The testing set has $2,228$ probe images ($702$ identities) and $17,661$ gallery images.

Both Cumulative Matching Characteristics (CMC) scores and mean average precision (mAP) are used for performance evaluation.

\vspace{1ex}\noindent\textbf{Baselines.}~Four base models are implemented. Specifically, we take ResNet-50~\cite{ResNet}, ResNeXt-50~\cite{resnext} and DenseNet-121~\cite{densenet} pretrained on ImageNet~\cite{imagenet} as the backbone models. The three networks are supervised by the cross-entropy loss, yielding three base models denoted as~\textbf{B1},~\textbf{B2} and~\textbf{B3}, respectively. Meanwhile, we also supervise ResNet-50~\cite{ResNet} with triplet loss~\cite{defense_triplet} and obtain the base model~\textbf{B4}.

All the models are trained using the Adam optimizer for $60$ epochs with a batch size of $32$. When testing, we extract the $L_2$ normalized activations from the networks before the loss layer as the image features.

\vspace{1ex}\noindent\textbf{State-of-the-art Methods.}~As there exists a huge number of re-ID algorithms~\cite{PolyMap,SCSP,Deepreid,DCIA,POP,iccv17dgm,eccv18race,yan2016person}, it is unrealistic to evaluate all of them. Here, we reproduce two representatives which achieve the latest state-of-the-art performances,~\ie,~Harmonious Attention CNN (HACNN)~\cite{HACNN} and Multi-task Attentional Network with Curriculum Sampling (Mancs)~\cite{Mancs}. Both of them employ attention mechanisms to address person misalignment. We follow the default settings correspondingly and report their performances as well as those of the four base models in Table~\ref{table_baselines}.

\begin{table}[!tb]
\small
\centering
\caption{Performance of the four base models and two state-of-the-art methods implemented in this work. The reproduced performances of~\cite{HACNN,Mancs} are slightly different from those reported in the original work.}
\vspace{-2ex}
\begin{tabular}{|l|*{2}{p{1.2cm}<{\centering}}|*{2}{p{1.2cm}<{\centering}}|}
\hline
\multirow{2}{*}{Methods} & \multicolumn{2}{c|}{Market-1501} & \multicolumn{2}{c|}{DukeMTMC-reID}  \\
\cline{2-3} \cline{4-5} 
                         & Rank-1 & mAP & Rank-1 & mAP \\
\hline
\hline
\textbf{B1} & 91.30 & 77.52 & 82.85 & 67.72 \\
\textbf{B2} & 91.44 & 78.21 & 83.03 & 67.85 \\
\textbf{B3} & 91.95 & 79.08 & 83.34 & 68.30 \\
\textbf{B4} & 84.29 & 67.86 & 76.57 & 57.31 \\
\cite{HACNN} & 90.56 & 75.28 & 80.70 & 64.44 \\
\cite{Mancs}  & 93.17 & 82.50 & 85.23 & 72.89 \\
\hline
\end{tabular}
\label{table_baselines}
\end{table}

\vspace{1ex}\noindent\textbf{Experimental Design.}~The design of  experiments involves various settings as described in Sec.~\ref{sec:methodology}. If not specified otherwise, we use the Euclidean distance defined in Eq.~\eqref{eq:euclidean} as the metric and I-FGSM defined in Eq.~\eqref{eq:I-FGSM} with $\epsilon=5$ as the attack method and perform white-box non-targeted attacks on base model $\textbf{B1}$. For other parameters, we set $\alpha=1$ in Eq.~\eqref{eq:I-FGSM} and $\mu=1$ in Eq.~\eqref{eq:MI-FGSM}. The iteration number $n$ is set to $\min(\epsilon+4,1.25\epsilon)$ following~\cite{I_FGSM}.


\subsection{White-box and Black-box Attack} \label{sec:exp_white_black}
Adversarial metric attack is first evaluated with a single model. For each query class, we generate adversarial examples on the corresponding gallery set. Thus, \emph{an adversarial version of the gallery set} can be stored off-line and used for performance evaluation. The maximum magnitude of adversarial perturbation $\epsilon$ is set to $5$ on Market-1501 in Table~\ref{table:market_white_black_5} and on DukeMTMC-reID in Table~\ref{table:duke_white_black_5}, which are still imperceptible to human vision (examples shown in Fig.~\ref{fig:ill_perturbation}). Therein, we present the networks that we attack in rows and networks that we test on in columns.

\begin{table*}[!tb]
\small
\centering
\caption{The mAP comparison of white-box attack (in shadow) and black-box attack (others) when $\epsilon=\textbf{5}$ on the Market-1501 dataset. For each model, the performance with clean images is put in the parentheses. For each combination of settings, the worst performances are marked in bold.}
\vspace{-2ex}
\resizebox{.925\textwidth}{!}{
\begin{tabular}{|l|p{1.82cm}|*{4}{p{1.7cm}<{\centering}}|*{2}{p{2.2cm}<{\centering}}|}
\hline
Model & Attack & \textbf{B1} (77.52) & \textbf{B2} (78.21) & \textbf{B3} (79.08) & \textbf{B4} (67.86)  & \cite{HACNN}~(75.28) & \cite{Mancs}~(82.50) \\
\hline
\hline
\multirow{3}{*}{\textbf{B1}} & FGSM      & \mycolor{7.054} & 33.73 & 36.43 & 30.53  & 44.00 & 43.49 \\
                    & I-FGSM    & \mycolor{\textbf{0.367}} & 25.12 & 29.42 & 24.11  & 43.51 & 34.68 \\
                    & MI-FGSM   & \mycolor{0.757} & \textbf{22.18} & \textbf{25.53} & \textbf{21.43}  & \textbf{37.98} & \textbf{30.90} \\
\hline
\hline
\multirow{3}{*}{\textbf{B2}} & FGSM      & 35.83 & \mycolor{10.47} & 42.71 & 37.24  & 48.45 & 51.23 \\
                    & I-FGSM    & 26.87 & \mycolor{\textbf{0.458}} & 35.96 & 32.91  & 48.06 & 45.69 \\
                    & MI-FGSM   & \textbf{23.01} & \mycolor{0.960} & \textbf{30.77} & \textbf{28.47}  & \textbf{41.84} & \textbf{39.80} \\
\hline
\hline
\multirow{3}{*}{\textbf{B3}} & FGSM      & 32.84 & 36.89 & \mycolor{9.178} & 33.91 & 44.40 & 45.95 \\
                    & I-FGSM    & 24.72 & 28.89 & \mycolor{\textbf{0.519}} & 29.26  & 43.99 & 39.49 \\
                    & MI-FGSM   & \textbf{22.29} & \textbf{26.13} & \mycolor{1.022} & \textbf{26.26}  & \textbf{38.92} & \textbf{35.31} \\
\hline
\hline
\multirow{3}{*}{\textbf{B4}} & FGSM     & 41.17 & 47.13 & 48.23 & \mycolor{4.320} & 51.10 & 51.87 \\
                    & I-FGSM   & 38.68 & 47.08 & 48.89 & \mycolor{\textbf{0.211}} & 54.72 & 50.89 \\
                    & MI-FGSM  & \textbf{32.31} & \textbf{39.58} & \textbf{41.47} & \mycolor{0.430} & \textbf{47.57} & \textbf{43.16} \\
\hline
\end{tabular}
}
\label{table:market_white_black_5}
\end{table*}

\begin{table*}[!tb]
\vspace{-1ex}
\small
\centering
\caption{The mAP comparison of white-box attack (in shadow) and black-box attack (others) when $\epsilon=\textbf{5}$ on the DukeMTMC-reID dataset. For each model, the performance with clean images is put in the parentheses. For each combination of settings, the worst performances are marked in bold.}
\vspace{-2ex}
\resizebox{.925\textwidth}{!}{
\begin{tabular}{|l|p{1.82cm}|*{4}{p{1.7cm}<{\centering}}|*{2}{p{2.2cm}<{\centering}}|}
\hline
Model & Attack & \textbf{B1} (67.72) & \textbf{B2} (67.85) & \textbf{B3} (68.30) & \textbf{B4} (57.31)  & \cite{HACNN}~(64.44) & \cite{Mancs}~(72.89) \\
\hline
\hline
\multirow{3}{*}{\textbf{B1}} & FGSM      & \mycolor{4.469} & 27.15 & 31.05 & 21.27 & 39.50 & 36.63 \\
                    & I-FGSM    & \mycolor{\textbf{0.178}} & 22.65 & 28.37 & 17.96 & 41.71 & 32.12 \\
                    & MI-FGSM   & \mycolor{0.315} & \textbf{17.16} & \textbf{22.01} & \textbf{14.08} & \textbf{33.96} & \textbf{24.74} \\
\hline
\hline
\multirow{3}{*}{\textbf{B2}} & FGSM & 27.87 & \mycolor{5.775} & 34.28 & 27.42 & 41.42 & 42.87 \\
 & I-FGSM & 24.58 & \mycolor{\textbf{0.159}} & 32.44 & 26.50 & 43.59 & 42.39 \\
 & MI-FGSM & \textbf{18.47} & \mycolor{0.342} & \textbf{25.39} & \textbf{20.58} & \textbf{36.09} & \textbf{33.37} \\
\hline
\hline
\multirow{3}{*}{\textbf{B3}} & FGSM & 26.54 & 28.36 & \mycolor{5.223} & 25.15 & 38.32 & 39.30 \\
 & I-FGSM & 22.43 & 23.96 & \mycolor{\textbf{0.191}} & 23.30 & 39.98 & 37.11 \\
 & MI-FGSM & \textbf{18.12} & \textbf{19.28} & \mycolor{0.387} & \textbf{18.93} & \textbf{33.29} & \textbf{29.85} \\
\hline
\hline
\multirow{3}{*}{\textbf{B4}} & FGSM  & 32.02 & 38.37 & 40.77 & \mycolor{2.071} & 45.52 & 43.70 \\
 & I-FGSM & 33.93 & 42.06 & 44.93 & \mycolor{\textbf{0.074}} & 50.61 & 47.34 \\
 & MI-FGSM & \textbf{24.99} & \textbf{32.34} & \textbf{35.32} & \mycolor{0.137} & \textbf{42.71} & \textbf{36.57} \\
\hline
\end{tabular}
}
\label{table:duke_white_black_5}
\end{table*}

At first glance, one can clearly observe the adversarial effect of different metrics. For instance, the performance of~\textbf{B1} decreases sharply from mAP $77.52$ to $0.367$ in white-box attack on Market-1501.~\myTextColor{This accuracy degradation is supported by an increase of mean distances with true positives from $0.80$ to $1.42$. Meanwhile, if directly performing classification attack via I-FGSM, the mAP achieved by~\textbf{B1} is $52.93$, which means that the distance metric is not effectively corrupted.}
On DukeMTMC-reID, the performance of~\textbf{B1} drops from $67.72$ to $0.178$ in white-box attack, and to $18.12$ in black-box attack. The state-of-the-art methods HACNN~\cite{HACNN} and Mancs~\cite{Mancs} are subjected to a dramatic performance decrease from mAP $75.28$ to $37.98$ and from $82.50$ to $30.90$, respectively on Market-1501\footnote{The analyses of rank-1 accuracy are in the supplementary material.}.
 

Second, the performance of white-box attack is much lower than that of black-box attack. It is easy to understand as the attack methods can generate adversarial examples that overfit the attacked model. Among the three attack methods, I-FGSM~\cite{I_FGSM} delivers the strongest white-box attacks. Comparatively, MI-FGSM~\cite{MI_FGSM} is the most capable of learning adversarial examples for black-box attack. This observation is consistent across different base models, different state-of-the-art methods, different magnitudes of adversarial perturbation\footnote{The results when $\epsilon=10$ are in the supplementary material.} and different datasets. This conclusion is somehow contrary to that drawn by classification attack~\cite{kurakin2016adversarial}, where non-iterative algorithms like FGSM~\cite{FGSM} can generally generalize better. In summary, we suggest integrating iteration-based attack methods for adversarial metric attack as they have a higher attack rate.

Moreover, HACNN~\cite{HACNN} and Mancs~\cite{Mancs} are more robust to adversarial examples compared with the four base models. When attacked by the same set of adversarial examples, they outperform baselines by a large margin, although Table~\ref{table_baselines} shows that they only achieve comparable or even worse performances with clean images. For instance in Table~\ref{table:market_white_black_5}, when attacking $\textbf{B1}$ using MI-FGSM in black-box setting, the best mAP achieved by the baselines is $25.53$ on Market-1501. In comparison, HACNN reports an mAP of $37.98$ and Mancs reports an mAP of $30.90$. A possible reason is that they both have more sophisticated modules and computational mechanisms,~\eg,~attention selection. However, it remains unclear and needs to be investigated in the future which kind of modules are robust and why they manifest robustness to adversary.

\begin{figure}[tb]
\centering
\includegraphics[width = 0.4\textwidth]{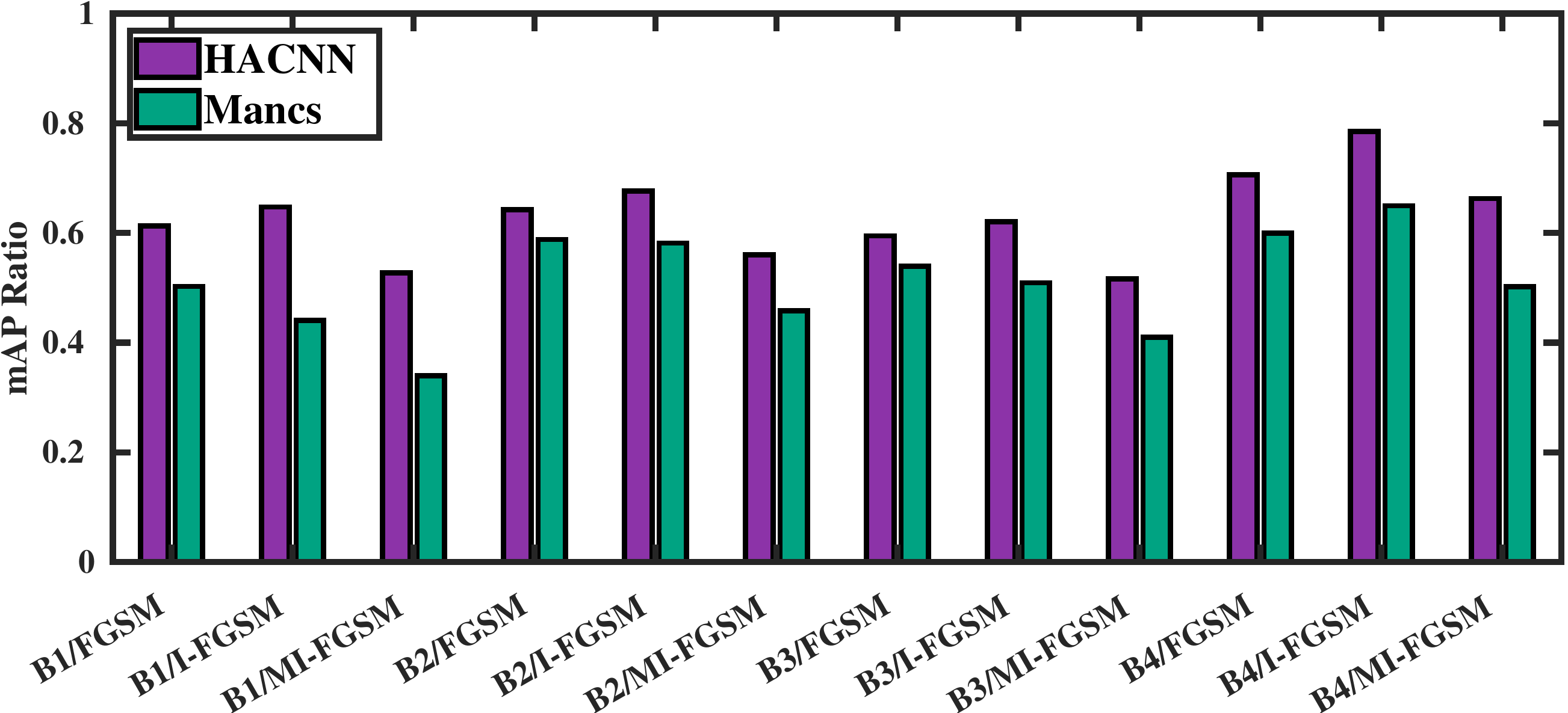}
\vspace{-2ex}
\caption{The ratio of mAP on adversarial examples to that on clean images on the DukeMTMC-reID dataset.}
\label{fig:ratio}
\vspace{-2ex}
\end{figure}

At last, the robustness of HACNN~\cite{HACNN} and Mancs~\cite{Mancs} to adversary are also quite different. In most adversarial settings, HACNN outperforms Mancs remarkably, revealing that it is less vulnerable to adversary. Only when attacking $\textbf{B2}$ or $\textbf{B3}$ using FGSM on DukeMTMC-reID, Mancs seems to be better than HACNN (mAP $42.87$ \emph{vs.} $41.42$ and $39.30$ \emph{vs.} $38.32$). However, it should be emphasized that the baseline performance of HACNN is much worse than that of Mancs with clean images as presented in Table~\ref{table_baselines} (mAP $75.28$ \emph{vs.} $82.50$ on Market-1501 dataset and mAP $64.44$ \emph{vs.} $72.89$ on DukeMTMC-reID). To eliminate the influence of the differences in baseline performance, we adopt a relative measurement of accuracy using the mAP ratio,~\ie,~the ratio of mAP on adversarial examples to that on clean images. A large mAP ratio indicates that the performance decrease is smaller, thus the model is more robust to adversary. We compare the mAP ratio of HACNN and Mancs in Fig.~\ref{fig:ratio}. As shown, HACNN consistently achieves a higher mAP ratio than Mancs in the adversarial settings. 

From another point of view, achieving better performances on benchmark datasets does not necessarily mean that the algorithm has better generalization capacity. Therefore, it would be helpful to evaluate re-ID algorithms under the same adversarial settings to justify the potential of deploying them in real environments. Meanwhile, the proposed adversarial metric attack is not limited to the baselines summarized in Table~\ref{table_baselines}. Instead, according to our experiments that are not reported here, it is applicable to a wide spectrum of metric losses (such as contrastive loss~\cite{contrastive} and quadruplet loss~\cite{quadruplet}) and other state-of-the-art methods (such as CAMA~\cite{CAMA}).



\subsection{Single-model and Multi-model Attack}
As shown in Sec.~\ref{sec:exp_white_black}, black-box attacks yield much higher mAP than white-box attacks, which means that the generated adversarial examples do not transfer well to other models for testing. Attacking multiple models simultaneously can be helpful to improve the transferability.

\begin{table*}[!tb]
\vspace{-1ex}
\small
\centering
\caption{The mAP comparison of multi-model attack (white-box in shadow) when $\epsilon=\textbf{5}$. The symbol ``-" indicates the name of the hold-out base model. For each combination of settings, the worst performances are marked in bold.}
\vspace{-2ex}
\resizebox{.925\textwidth}{!}{
\begin{tabular}{|l|l|cccc|cccc|}
\hline
\multirow{2}{*}{Model} & \multirow{2}{*}{Attack} & \multicolumn{4}{c|}{Market-1501} & \multicolumn{4}{c|}{DukeMTMC-reID}  \\
\cline{3-10}
& & Ensemb. & Hold-out & HACNN~\cite{HACNN} & Mancs~\cite{Mancs}  & Ensemb.& Hold-out & HACNN~\cite{HACNN} & Mancs~\cite{Mancs} \\
\hline
\hline
\multirow{3}{*}{-\textbf{B1}} & FGSM     & \mycolor{20.61} & 26.52 & 40.68 & 39.37 & \mycolor{13.65} & 20.52 & 35.11 & 32.26  \\
                     & I-FGSM   & \mycolor{\textbf{3.839}} & \textbf{14.55} & 35.62 & 26.35 & \mycolor{\textbf{2.058}} & 12.72 & 32.11 & 23.48  \\
                     & MI-FGSM  & \mycolor{5.900} & 14.94 & \textbf{33.15} & \textbf{25.88} & \mycolor{3.213} & \textbf{11.78} & \textbf{28.23} & \textbf{20.81}  \\
\hline
\hline
\multirow{3}{*}{-\textbf{B2}} & FGSM     & \mycolor{19.76} & 29.12 & 39.43 & 36.89 & \mycolor{13.32} & 22.83 & 34.33 & 30.21  \\
                     & I-FGSM   & \mycolor{\textbf{3.801}} & \textbf{17.03} & 34.40 & \textbf{22.87}  & \mycolor{\textbf{2.019}} & 14.11 & 31.31 & 20.04  \\
                     & MI-FGSM  & \mycolor{5.840} & 17.48 & \textbf{32.21} & 23.04 & \mycolor{3.125} & \textbf{13.18} & \textbf{27.60} & \textbf{18.18}  \\
\hline
\hline
\multirow{3}{*}{-\textbf{B3}} & FGSM     & \mycolor{20.64} & 32.62 & 40.62 & 38.64  & \mycolor{13.99} & 26.88 & 35.34 & 31.26  \\
                     & I-FGSM   & \mycolor{\textbf{3.839}} & 21.20 & 35.58 & 24.75 & \mycolor{\textbf{2.089}} & 19.46 & 32.80 & 21.42  \\
                     & MI-FGSM  & \mycolor{5.905} & \textbf{20.73} & \textbf{32.89} & \textbf{24.44} & \mycolor{3.265} & \textbf{17.44} & \textbf{28.45} & \textbf{18.86}  \\
\hline
\hline
\multirow{3}{*}{-\textbf{B4}} & FGSM     & \mycolor{21.37} & 26.07 & 38.15 & 36.61 & \mycolor{13.96} & 17.72 & 32.37 & 29.28  \\
                     & I-FGSM   & \mycolor{\textbf{4.521}} & \textbf{16.47} & 32.64 & \textbf{22.27} & \mycolor{\textbf{2.483}} & 11.65 & 28.99 & 18.52  \\
                     & MI-FGSM  & \mycolor{6.847} & 16.66 & \textbf{30.45} & 22.75 & \mycolor{3.693} & \textbf{10.93} & \textbf{25.46} & \textbf{17.14}  \\
\hline
\end{tabular}
}
\label{table:ensemble}
\vspace{-2ex}
\end{table*}

To achieve this, we perform adversarial metric attack on an ensemble of three out of the four base models. Then, the evaluation is done on the ensembled network and the hold-out network. Note that in this case, attacks on the ``ensembled network" correspond to white-box attacks as the base models in the ensemble have been seen by the attacker during adversarial metric attack. In contrast, attacks on the ``hold-out network" correspond to black-box attacks as this network is not used to generate adversarial examples.

We list the performances of multi-model attacks in Table~\ref{table:ensemble}. As indicated clearly, the identification rate of black-box attacks continues to degenerate. For example, Table~\ref{table:market_white_black_5} shows that the worst performance of $\textbf{B1}$ is mAP $22.29$ when attacking the single model $\textbf{B3}$ via MI-FGSM on Market-1501. Under the same adversarial setting, the performance of $\textbf{B1}$ becomes $14.94$ when attacking an ensemble of $\textbf{B2}$, $\textbf{B3}$ and $\textbf{B4}$. When attacking multiple models, the lowest mAP of HACNN~\cite{HACNN} is merely $30.45$ on the Market-1501 dataset, a sharp decrease of $7.53$ from $37.98$ as reported in Table~\ref{table:market_white_black_5} under the same adversarial settings.

\subsection{Targeted and Non-targeted Attack}
\begin{figure*}[tb]
\centering
\subfigure[]
{
\begin{minipage}[tb]{0.4\textwidth}
\includegraphics[width = 1\textwidth]{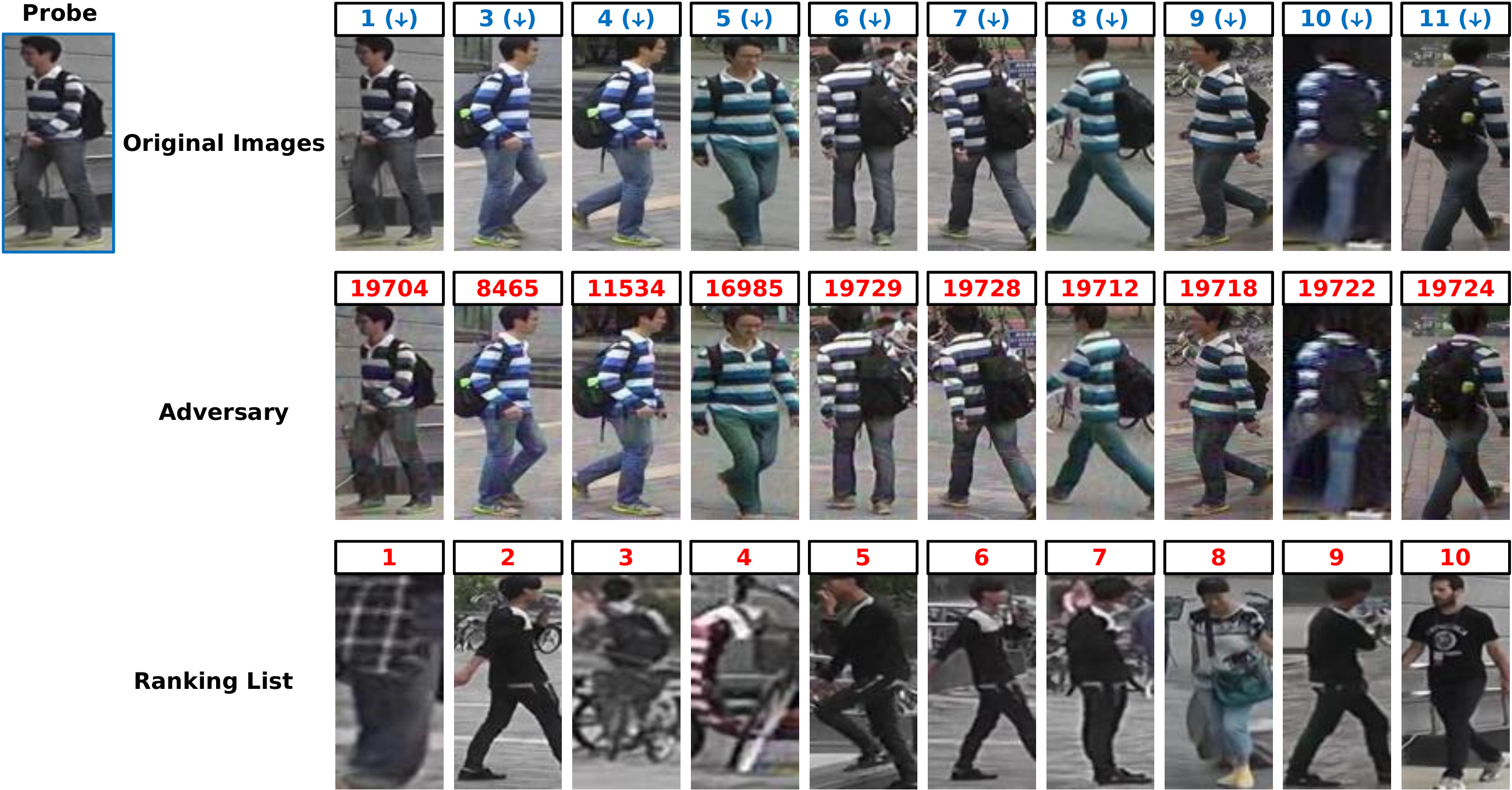}
\end{minipage}
\label{fig:non-targeted}
}
\subfigure[]
{
\begin{minipage}[tb]{0.4\textwidth}
\includegraphics[width = 1\textwidth]{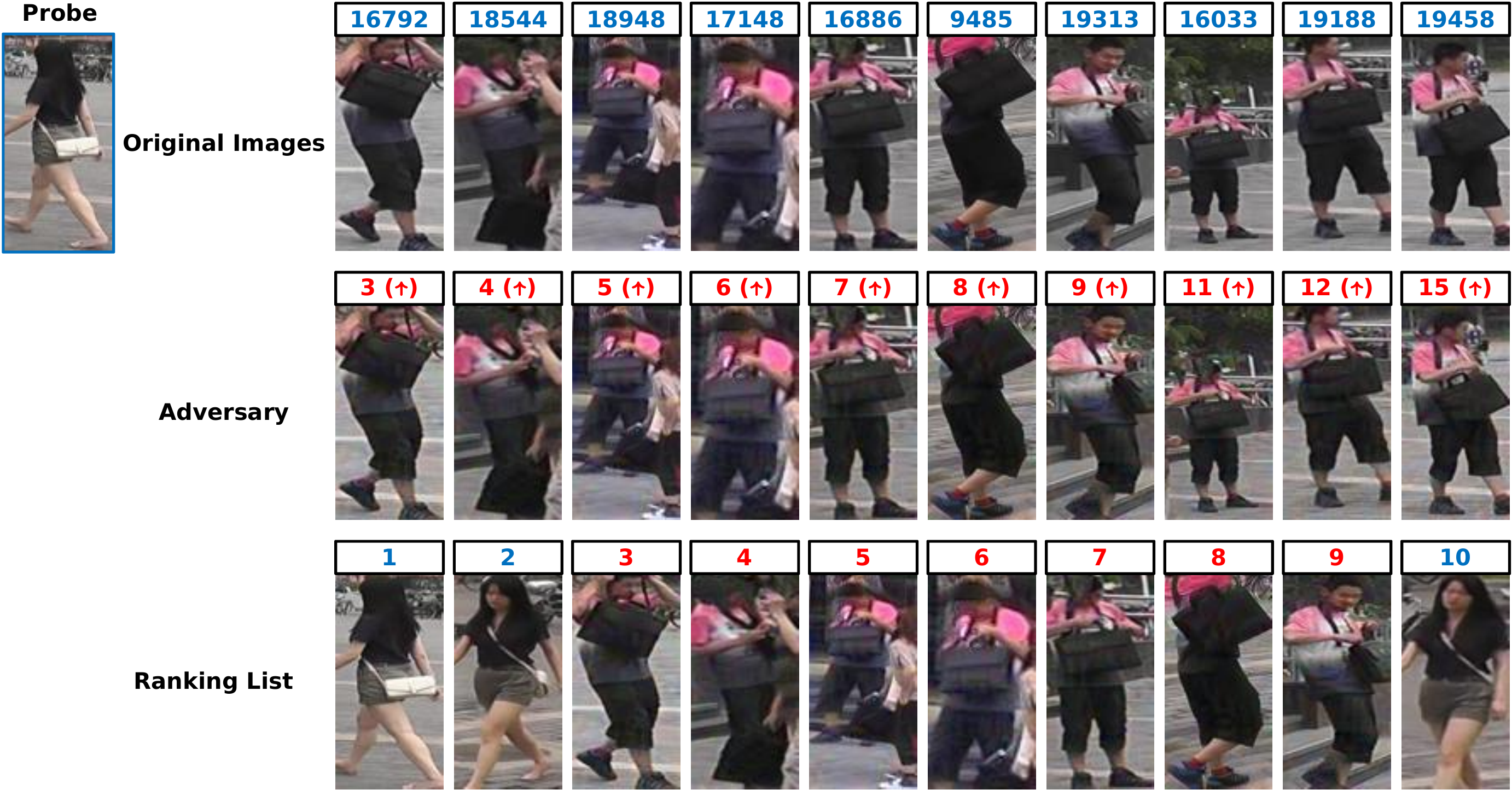}
\end{minipage}
\label{fig:targeted}
}
\vspace{-2ex}
\caption{Two representative ranking lists of two probe images for non-targeted attack (a) and targeted attack (b). We mark the ranking position of each gallery image on its top and do not elaborately exclude the distractor images and those captured by the same camera as the probe. The gallery images with proper ranking positions (\ie,~true positives and true negatives) are marked in blue, otherwise in red.}
\label{fig:tar_no_tar}
\vspace{-2ex}
\end{figure*}

From Fig.~\ref{fig:tar_no_tar}, one can clearly observe the different effects of non-targeted and targeted attacks.
The goal of non-targeted metric attack is to maximize the distances (minimize the similarities) between a given probe and adversarial gallery images. Consequently, true positives are pushed down in the ranking list as shown in the first two rows of Fig.~\ref{fig:non-targeted}. However, it is indeterminable beforehand what the top-ranked images will be and to which probe the adversary will be similar as shown in the third row. 
In comparison, a targeted metric attack tries to minimize the distances between the given probe and the adversarial gallery images. Therefore, we find a large portion of adversarial images in top-ranked candidates in the third row of Fig.~\ref{fig:targeted}. And it is surprising to see that the metric is so easy to be fooled, which incorrectly retrieves male person images when indexing a female person image.
For real applications in video surveillance, the non-targeted metric attack prevents the system from correctly retrieving desired results, while the targeted metric attack deliberately tricks the system into retrieving person images of a wrong identity.

\begin{figure}[tb]
\centering
\subfigure[]
{
\begin{minipage}[tb]{0.2\textwidth}
\includegraphics[width = 1\textwidth]{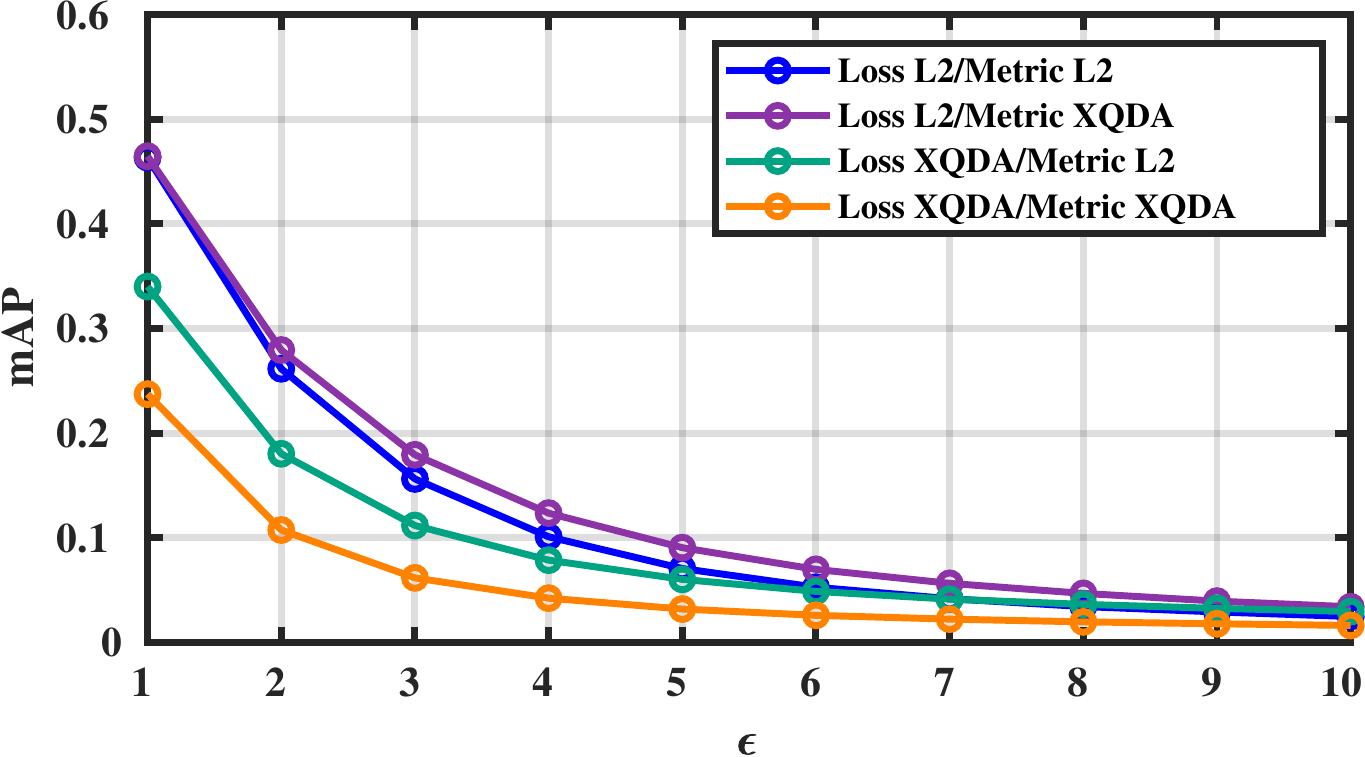}
\end{minipage}
\label{fig:fgsm}
}
\subfigure[]
{
\begin{minipage}[tb]{0.2\textwidth}
\includegraphics[width = 1\textwidth]{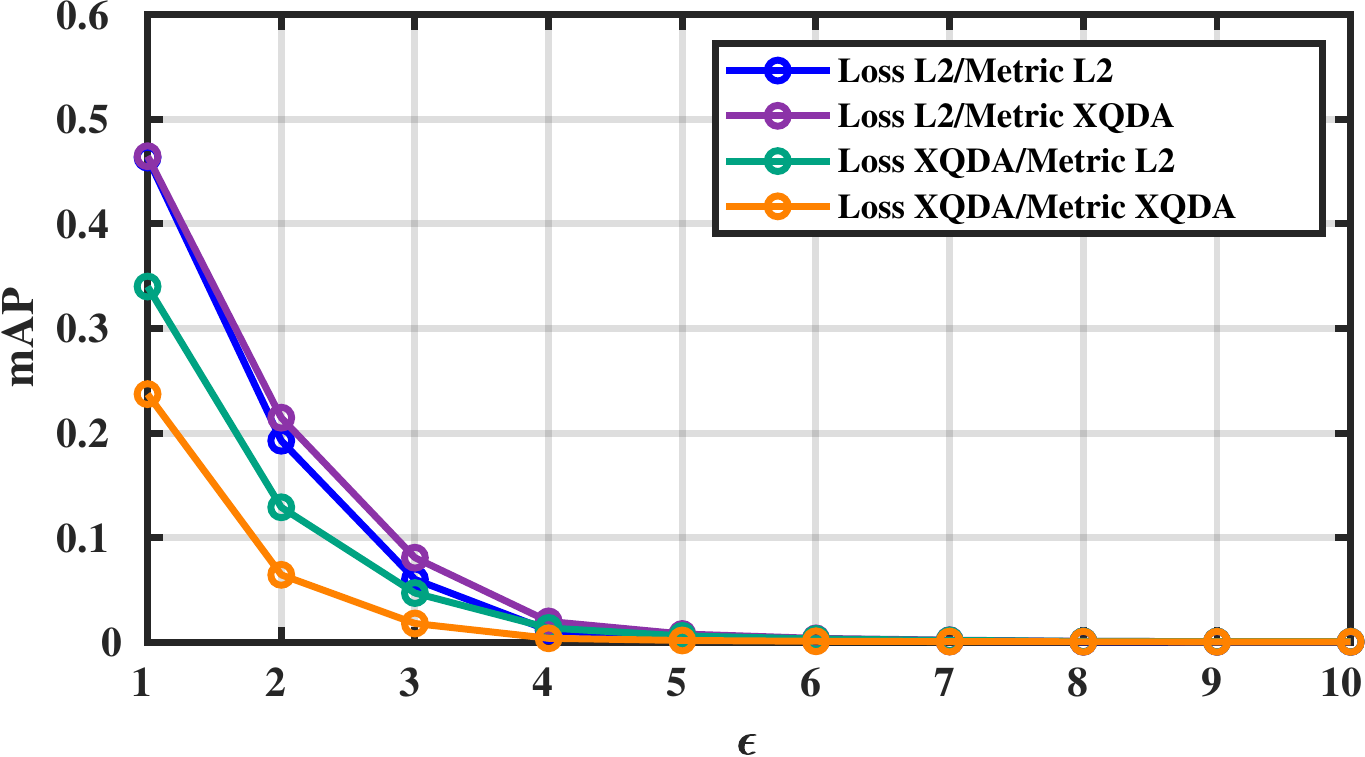}
\end{minipage}
\label{fig:ifgsm}
}
\vspace{-2ex}
\caption{The mAP comparison of FGSM (a) and I-FGSM (b) by varying the maximum magnitude of adversarial perturbation $\epsilon$ and a selection of distance metric. In the legend, the part before symbol ``/" denotes the metric loss used for metric attack and the part after ``/" denotes the metric used to evaluate the performance.}
\label{fig:xqda}
\vspace{-2ex}
\end{figure}

\begin{table*}[!tb]
\small
\centering
\caption{The mAP comparison between normally trained models (denoted by \#N) and metric-preserving models (denoted by \#M) on the Market-1501 dataset. \#I means the relative improvement.}
\vspace{-2ex}
\resizebox{.925\textwidth}{!}{
\begin{tabular}{|l|ccc|ccc|ccc|ccc|}
\hline
\multirow{2}{*}{Gallery} & \multicolumn{3}{c|}{\textbf{B1}} & \multicolumn{3}{c|}{\textbf{B2}} & \multicolumn{3}{c|}{\textbf{B3}} & \multicolumn{3}{c|}{\textbf{B4}}\\
\cline{2-13}
     & \#N & \#M & \#I & \#N & \#M & \#I& \#N & \#M & \#I & \#N & \#M & \#I \\
\hline
\hline
Original           & 77.52 & 76.69 & -1.07\% & 78.21 & 76.74 & -1.87\% & 79.08 & 77.25 & -2.31\% & 67.86 & 59.87 & -11.7\% \\
Adv. (\textbf{B1}) & 0.367 & 74.14 & +2.0e4\% & 25.12 & 70.54 & +180\% & 29.42 & 72.45 & +146\% & 24.11 & 53.58 & +122\%  \\
Adv. (\textbf{B2}) & 26.87 & 72.64 & +170\% & 0.458 & 76.23 & +1.6e4\% & 35.96 & 72.82 & +102\% & 32.91 & 52.81 & +60.4\% \\
Adv. (\textbf{B3}) & 24.72 & 70.46 & +185\% & 28.89 & 68.67 & +137\% & 0.519 & 76.93 & +1.4e4\% & 29.26 & 51.03 & +74.4\% \\
Adv. (\textbf{B4}) & 38.68 & 72.65 & +87.8\% & 47.08 & 72.14 & +53.2\% & 48.89 & 73.41 & +50.1\% & 0.211 & 57.45 & +2.7e4\% \\
\hline
\end{tabular}
}
\label{table:defense}
\vspace{-2ex}
\end{table*}

\subsection{Euclidean and Mahalanobis Metric}
Fig.~\ref{fig:xqda} plots the mAP comparison of FGSM and I-FGSM by varying the maximum magnitude of adversarial perturbation $\epsilon$ and a selection of distance metric on Market-1501. 

Within our framework, distance metrics can be used in two phases, that is, the one used to perform adversarial metric attack and the one used to evaluate the performance. For the Mahalanobis distance, we use a representative called Cross-view Quadratic Discriminant Analysis (XQDA)~\cite{XQDA}. Unfortunately, by integrating metric learning with deep features, we do not observe an improvement of baseline performance, despite the fact that metric learning is extensively proven to be compatible with non-deep features (\eg,~LOMO~\cite{XQDA}, GOG~\cite{GOG}). We obtain a rank-1 accuracy of $89.73$ and an mAP of $75.86$ using XQDA, lower than the rank-1 accuracy of $91.30$ and mAP of $77.52$ achieved by the Euclidean distance reported in Table~\ref{table_baselines}.

From Fig.~\ref{fig:xqda}, it is unsurprising to observe that the performance of different metric combinations decreases quickly as the maximum magnitude of adversarial perturbation $\epsilon$ increases. We also note that the iteration-based attack methods such as I-FGSM and MI-FGSM can severely mislead the distance metric with $5$-pixel perturbations.

Second, we observe an interesting phenomenon which is consistent with different attack methods. When attacking the Euclidean distance and testing with XQDA, the performance is better than the setting where attacking and testing are both carried out with the Euclidean distance. This is also the case when we attack XQDA and test with the Euclidean distance. In other words, it is beneficial to adversarial metric defense if we use different metrics for metric attack and performance evaluation. From another perspective, it can be interpreted by the conclusion drawn in Sec.~\ref{sec:exp_white_black},~\ie,~we can take the change of metrics as a kind of black-box attack. In other words, we are using adversarial examples generated with a model using a certain distance metric to test another model which differs from the original model in its choice of distance metric.

\subsection{Evaluating Adversarial Metric Defense}
In Table~\ref{table:defense}, we evaluate metric defense by comparing the performance of normally trained models with metric-preserving models on Market-1501. When testing the original clean gallery set, a slight performance decrease, generally smaller than $10\%$, is observed after using metric-preserving models. However, when purely testing the adversarial version of gallery images, the performance is significantly improved. For instance, when attacking $\textbf{B3}$ and testing on $\textbf{B1}$, the performance is originally $24.72$, then improved to $70.46$ with a relative improvement of $185\%$. In real video surveillance, it can improve the robustness of re-ID systems by deploying metric-preserving models.

\section{Conclusion} \label{sec:con}
In this work, we have studied the adversarial effects in person re-identification (re-ID). By observing that most existing works on adversarial examples only perform classification attacks, we propose the adversarial metric attack as a parallel methodology to be used in  metric analysis. 

By performing metric attack, adversarial examples can be easily generated for person re-identification. The latest state-of-the-art re-ID algorithms suffer a dramatic performance drop when they are attacked by the adversarial examples generated in this work, exposing the potential security issue of deploying re-ID algorithms in real video surveillance systems. To facilitate the development of metric attack in person re-identification, we have benchmarked and introduced various adversarial settings, including white-box and black-box attack, targeted and non-targeted attack, single-model and multi-model attack,~\etc. Extensive experiments on two large scale re-ID datasets have reached some useful conclusions, which can be a helpful reference for future works. Moreover, benefiting from adversarial metric attack, we present an early attempt of training metric-preserving networks, which we believe, still requires further exploration in re-ID. The code is publicly available to facilitate the development of adversarial metric attack and defense of re-ID systems.



{\small
\bibliography{example_paper}
\bibliographystyle{IEEEtran}
}

\end{document}